\documentclass[conference]{IEEEtran}
\IEEEoverridecommandlockouts
\usepackage{cite}

\usepackage{textcomp}
\usepackage{xcolor}
\def\BibTeX{{\rm B\kern-.05em{\sc i\kern-.025em b}\kern-.08em
    T\kern-.1667em\lower.7ex\hbox{E}\kern-.125emX}}

\usepackage{amsmath,amssymb,amsfonts}
\usepackage{algorithmic, algorithm}

\usepackage{multicol}
\usepackage{subcaption}
\usepackage{times}  
\usepackage{helvet} 
\usepackage{courier}  
\usepackage[hyphens]{url}  
\usepackage{graphicx} 
\usepackage{caption} 
\usepackage{color}
\usepackage{soul}
\usepackage[super]{nth}
\usepackage{url}
\usepackage[normalem]{ulem}

\begin{document}

\title{TRUST XAI: Model-Agnostic Explanations for AI With a Case Study on IIoT Security}

\author{Maede~Zolanvari,~\IEEEmembership{Student Member,~IEEE,}
		Zebo~Yang,~\IEEEmembership{Student Member,~IEEE,}
		Khaled~Khan,~\IEEEmembership{Member,~IEEE,} \\
        Raj~Jain,~\IEEEmembership{Life~Fellow,~IEEE,}
        and~Nader~Meskin,~\IEEEmembership{Senior~Member,~IEEE}
\thanks{Manuscript is submitted on May 1, 2021 to IEEE Internet of Things Journal Special Issue on Artificial Intelligence-Based Systems for Industrial
Internet of Things and Industry 4.0. This work was supported in part by the Qatar National Research Fund (QNRF) under Grant NPRP10-0206-170360; and in part by NSF under Grant CNS-1718929 (Corresponding author: Maede Zolanvari.)}
\thanks{Maede Zolanvari, Zebo Yang, and Raj Jain are with Department of Computer Science and Engineering, Washington University, St. Louis, MO USA (e-mail: maede.zolanvari@wustl.edu; zebo@wustl.edu; jain@wustl.edu).}
\thanks{Khaled Khan and Nader Meskin are with Department of Computer Science and Engineering, Qatar University, Doha, Qatar (e-mail: k.khan@qu.edu.qa; nader.meskin@qu.edu.qa).}}

\maketitle
\thispagestyle{plain}
\pagestyle{plain}

\begin{abstract}
Despite AI's significant growth, its ``black box" nature creates challenges in generating adequate trust. Thus, it is seldom utilized as a standalone unit in IoT high-risk applications, such as critical industrial infrastructures, medical systems, and financial applications, etc. Explainable AI (XAI) has emerged to help with this problem. However, designing appropriately fast and accurate XAI is still challenging, especially in numerical applications. Here, we propose a universal XAI model named Transparency Relying Upon Statistical Theory (TRUST), which is model-agnostic, high-performing, and suitable for numerical applications. Simply put, TRUST XAI models the statistical behavior of the AI's outputs in an AI-based system. Factor analysis is used to transform the input features into a new set of latent variables. We use mutual information to rank these variables and pick only the most influential ones on the AI’s outputs and call them ``representatives" of the classes. Then we use multi-modal Gaussian distributions to determine the likelihood of any new sample belonging to each class. We demonstrate the effectiveness of TRUST in a case study on cybersecurity of the industrial Internet of things (IIoT) using three different cybersecurity datasets. As IIoT is a prominent application that deals with numerical data. The results show that TRUST XAI provides explanations for new random samples with an average success rate of 98\%. Compared with LIME, a popular XAI model, TRUST is shown to be superior in the context of performance, speed, and the method of explainability. In the end, we also show how TRUST is explained to the user.
\end{abstract}


\section{Introduction}
\IEEEPARstart{T}{he} impact of artificial intelligence (AI) on today's technological advancements is undeniable. Its applications range from daily life events such as medical decision support to fundamental demands such as cybersecurity. Despite the popularity of AI, it is limited by its current inability to build trust. Researchers and industrial leaders have a hard time explaining the decisions that sophisticated AI algorithms come up with because they (as AI users) cannot fully understand why and how these ``black boxes" make their decisions.

Trusting AI blindly impacts its applicability and legitimacy. Based on Gartner's predictions, 85\% of the AI projects until 2022 will produce inaccurate outcomes resulting from the organization's limited knowledge of the deployed AI and its behavior \cite{Gartner2018}. Also, the increasing growth in implementing AI has raised concerns of 91\% of cybersecurity professionals about cyberattacks using AI \cite{Webroot2017}. In the medical field, dedicated AI algorithms are able to precisely (sometimes even more accurately than a human practitioner) detect rare diseases \cite{Bartoletti2019}. However, due to the lack of AI's decision transparency and explainability, the results are not considered trustworthy enough to be utilized in practice \cite{Holzinger2017}. A global survey shows that more than 67\% of the business leaders do not trust AI, and they believe AI's ambiguity will have negative impacts on their business in the next five years \cite{Oxborough2018}.

Moreover, the Internet of Things (IoT), as an integral part of both personal lives and industrial environments, has profited from AI for AI-powered data analysis. It allows connections, interactions, and data exchange among machines and devices for comprehensive functionality and higher efficiency without requiring any human in the loop. This automation requires a tremendous amount of trust while utilizing AI. This fact emphasizes why incorporating Explainable AI (XAI) in an AI-based IoT system is essential.

Designing self-explanatory models has gained much attention recently. XAI deals with psychology and cognitive science to provide transparent reasonings for AI's actions \cite{Miller2019}. Based on how the underlying AI is designed and the input types (image, text, voice, numerical data), the methods of explainability differ.

A proper XAI model should be integrated into the system to gain trust and transparency for AI-based systems. For clarity, in this paper, we call the main underlying AI model the \textit{primary AI model} and the explainable model that accompanies it, the \textit{XAI model}. The primary AI model is in charge of the main functionality of the system, such as classification, regression, recommendation, recognition, etc. The XAI model is in charge of providing transparency and explanation in the primary AI's behavior.

The primary AI and the XAI could be both in one model, which we can interpret the model's outcomes directly using the model itself. This is achievable by using interpretable models as the primary AI model (e.g., decision tree, linear regression, etc.). Even though this type of XAI might provide potentially more accurate explanations, it is limited by the non-reusability, and we can only count on the model providers \cite{Adadi2018}. Also, learning performance and explainability of learning models usually have an inverse relationship \cite{Holzinger2017}. Better learning performance comes with more sophistication in finding complex relationships in the data features. As such, it becomes more challenging for individuals to grasp its rationality. Therefore, in some applications, choosing interpretable models as the primary AI might yield a significantly degraded learning performance. Hence, surrogate explainers\footnote{This means developing a separate model working in parallel with the primary AI model to explain its behavior.} running in parallel along with the primary AI have become preferable in the explanation domain. They are independent and do not affect or put restrictions on the performance of the primary AI model. 

It is important to note that, in the XAI domain, it is not expected that the explanation would be in layman's terms. Most of the developed XAI models in the literature are just another interpretable AI or machine learning (ML) model that is easily understood. Therefore, it is assumed that the user is an expert in the internal workings of the learning model utilized as the XAI and knows the mathematics behind it. Here, we argue that statistics are significantly more verifiable and explainable than the AI or ML models used as the XAI. Therefore, we propose using statistical models to simplify the explanation process.

This paper introduces an XAI model that provides transparency, relying upon statistical theory (TRUST). TRUST XAI is a surrogate explainer that provides interpretability without sacrificing the performance of the primary model or imposing restrictions on it. The explainer does not depend on the type of the primary AI in any way; thus, it is entirely model-agnostic\footnote{This simply means the XAI does not care about the primary AI algorithm. All it requires is being able to probe the model with any arbitrary input and get an output from the ``black box".}. We believe that TRUST XAI will complement the future AI deployments as a standalone unit by providing transparent, independent, and reliable explanations.

While many papers in the XAI domain have focused on image-based data \cite{Adadi2018, Liu2019}, TRUST XAI applies to numerical data such as network data for IoT and security systems, which is critical but currently deficient in XAI.

\begin{figure}[t]
\centering
  \includegraphics[scale=0.6]{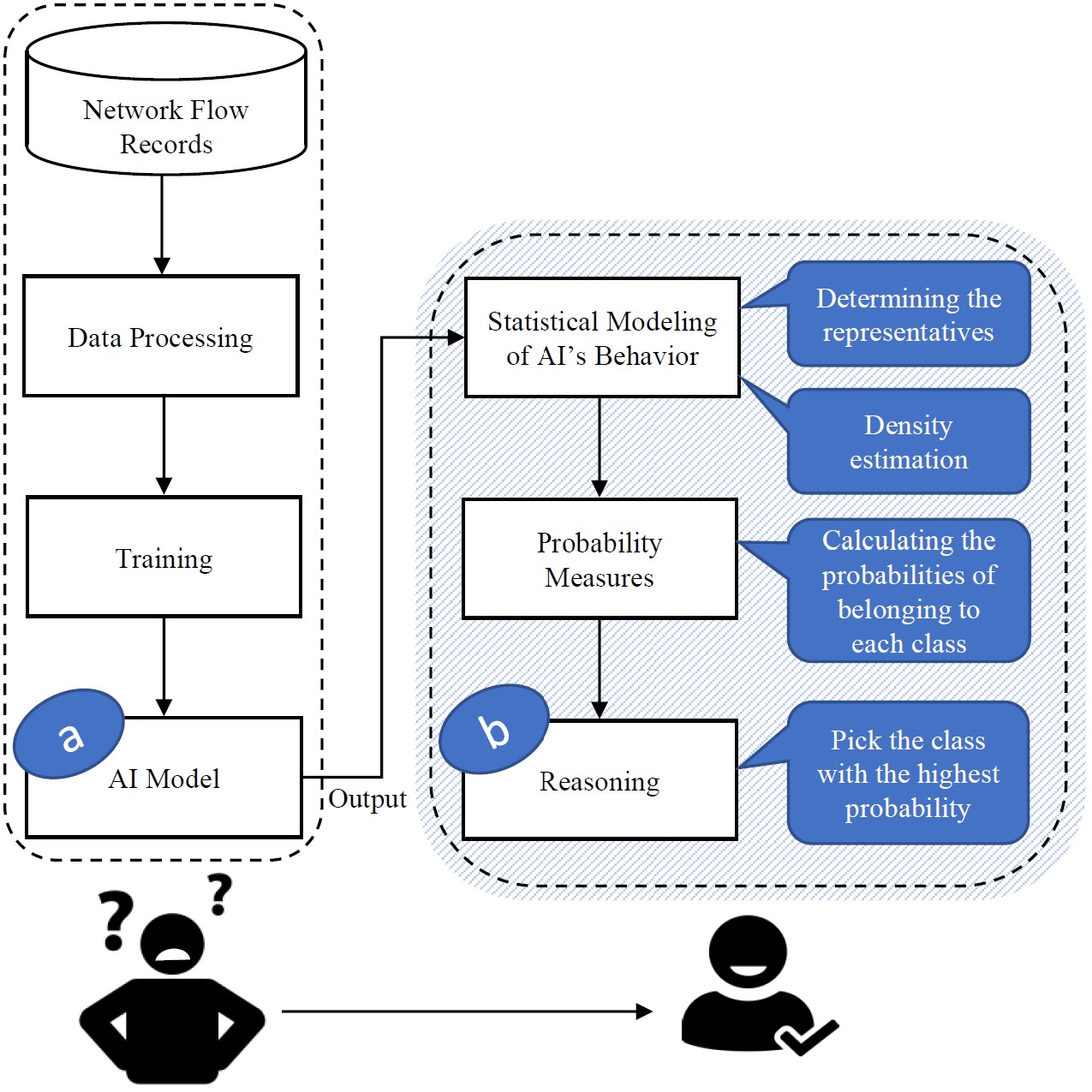}
  \caption{Integrated TRUST explainer in an AI-based system. While (a) can be very complex and non-interpretable, (b) should be as easily interpretable as possible for humans.}
  \label{XAI}
\end{figure}

The TRUST technique works solely with the feature sets and the outputs from the primary AI. If the primary AI is modeled as shown in Figure \ref{XAI} on the left side, TRUST XAI is represented by the three steps on the right using a background with diagonal stripes. The first step builds the core of the explainer, and it has to be calculated just once. Then, only the last two steps are needed for explaining any new data instances.

To evaluate and provide a practical instance for TRUST XAI, we use three datasets: 1) ``WUSTL-IIoT," a dataset generated from our prior work on our industrial IoT (IIoT) testbed \cite{Zolanvari2019}; 2) ``NSL-KDD," a public dataset for network-based IDSs from University of New Brunswick \cite{Tavallaee2009}; and 3) ``UNSW," another IDS dataset from the Cyber Range Lab of the Australian Centre for Cyber Security (ACCS) \cite{Moustafa2015}. We empirically prove that our proposed XAI model successfully explains the labels predicted by the primary AI in just a few milliseconds. Moreover, we have implemented and applied Local Interpretable Model-agnostic Explanations (LIME) on the three mentioned datasets for the sake of comparison. Our results show LIME is much slower than our model and is not practical for real-time applications or when dealing with a large number of samples.

The research contributions of this work are as follows: 
\begin{enumerate}
\item We propose a universal, accurate, and well-performing XAI model called TRUST XAI that requires no compromise in the selection or performance of the primary AI, while at the same time providing transparency in the model's results.
\item We show that TRUST XAI is very fast at reasoning the primary AI model's behavior on numerical data, making it one of the first XAI models suitable for real-time numerical applications.
\item Our case study introduces a unique perspective on the assessment of AI-based trustworthy systems for IIoT and Industry 4.0 \cite{Gilchrist2016}. 
\item Our IIoT security dataset would be released to support the research community for a more extensive and more diverse data collection in the emerging field of AI and XAI for IIoT.
\end{enumerate}

As data privacy and integrity are essential in all areas of IoT, cybersecurity is of paramount importance. Utilizing AI to provide a secure platform in this area has been proven as an effective method, especially in mission-critical applications such as IIoT \cite{Costa2019}, \cite{Alcaraz2019}, and \cite{Zolanvari2019}. Our case study on the security of IIoT indicates our belief in the importance of IIoT and Industry 4.0. However, TRUST XAI is universal and can be applied to any other applications with minor or zero modification.

The rest of the paper is organized as follows. The related research papers are explored in Section II. in Section III, we discussed the proposed TRUST model, including how we formalize the problem, what the assumptions are, and how we develop a proper model. Following that, to test TRUST, we test its performance through a case study in Section IV. How TRUST would interpret the results to the user is explained in Section V. Finally, Section V concludes the paper, while mentioning the limitations of TRUST and future work.

\section{Related Work}
The application of XAI in the IoT domain has emerged as a substantial promise for improving trust and transparency. Examples include industrial settings \cite{Gade2019}, \cite{Saad2019}, smart grids \cite{Siniosoglou2021}, 5G telecommunication \cite{Guo2020}, smart homes \cite{Magarino2019}, and healthcare \cite{Ahmad2018}, \cite{Tjoa2020}, along with XAI surveys covering different other areas, such as \cite{Adadi2018}, \cite{Guidotti2018}, \cite{Daglarli2021}, \cite{AfzaliSeresht2019}, \cite{Mathews2019}, and \cite{Arrieta2020}. XAI has been recognized as a necessary trend toward the next generation of AI-based systems, aiming to produce transparent models without affecting the primary AI's accuracy. The trade-offs between accuracy, interpretability, reusability, and performance have been continuously studied \cite{Adadi2018, Guidotti2018, Miller2019, Bartoletti2019}.

As discussed in \cite{Molnar2018}, feature importance is the most commonly utilized method of explainability. It is the basic technique to find the key features impacting the AI’s outcome. Feature selection is a common tool in data preprocessing \cite{Li2017}. It discovers important features that affect the AI's output in general across all classes. However, as seen in the following works, pinpointing these features alone cannot provide a baseline for evaluating or comparing the performance of the proposed XAI.

For example, in \cite{Wang2020}, the SHAP (SHapley Additive exPlanations) values, which is a feature importance technique originally developed by \cite{Lundberg2017}, are used to explain the predictions made by the primary AI model in network security. The results of the XAI model are presented as the list of features that are more indicative of the class of data. However, even the authors confirmed that SHAP is computationally heavy and not suitable for real-time applications. In \cite{Amarasinghe2018}, another feature importance technique, layer-wise relevance propagation (LRP), is developed and specialized for the explainability of deep neural network (DNN) models. It should potentially provide accurate explanations for DNN, but it cannot be generalized for other models.

In \cite{Sagi2020}, modeling a random forest with a decision tree has been proposed to add interpretability to the results. We know that while decision trees are highly explainable, random forest suffers from being complex and lack in interpretability. However, this technique is not model-agnostic since it is only applicable in tree-based learners. In \cite{Marino2018}, an adversarial technique is used as a means of interpretability. The method calculates the minimum adjustments to the important features' values of a misclassified sample until the primary AI correctly predicts it. However, the proposed method is semi-agnostic since it works only for the AI models with a defined gradient cross-entropy loss function. Additionally, no performance assessment is provided.

Finally, LIME, which is used as a \textit{benchmark XAI} in this paper, has been a popular technique in the XAI domain, which has many derivatives since its introduction in \cite{Ribeiro2016}, such as \cite{Zafar2019, Rabold2019}. It generates local surrogate models for every instance to explain its label. For each instance, a new dataset is synthesized consisting of samples with feature values drawn from a normal distribution with a mean and standard deviation corresponding to that instance. The primary AI then labels the synthesized samples. Also, they are assigned weights based on their distance from the instance of interest. Next, a simple interpretable AI or ML model (usually a linear regression model) is trained with the new synthetic dataset to explain the outcome of the primary AI for that specific instance. LIME is re-evaluated and compared with Anchors introduced in \cite{Ribeiro2018} by the same authors. Their results show that a linear explanation like LIME is preferred in cases where predictions are near the primary AI's decision boundary because linear explanations provide insight into the primary AI, while Anchors contributes by providing the coverage of the explanations with high accuracy (e.g., when the word "not" has a positive influence on the sentiment, such as "not bad"). Later in this paper, we compare the performance of LIME with our TRUST model and show how our model outperforms LIME in terms of speed and performance.

\section{Proposed TRUST Explainer}
The goal of TRUST XAI is to explain the rationale behind the data labeling of the primary AI model by applying statistical techniques to the AI's outputs. Table \ref{symbols} presents the symbols used in this paper. Symbols that are explained along with the equations or only used once are skipped from this table due to length considerations.

\begin{table}[t]
\centering\fontsize{9}{11}\selectfont
\caption{Symbol Table}
\begin{tabular}{|c|p{7cm}|}
\hline
\textbf{Symbol} & \textbf{Description} \\
\hline
$N$& Number of observations in the training set\\
\hline
$C$ & Total number of classes\\
\hline
$c$ & Index indicating class number, $c\in\{1,...,C\}$\\
\hline
$N_c$ & Number of observations classified by the primary AI as being in class $c$, $\sum_{c=1}^C N_c = N$\\
\hline
$K$& Number of features used in the primary AI model\\
\hline
$k$& Number of representatives used in the XAI model, $k\ll K$ \\
\hline
$i$& Index indicating the feature number $i\in\{1,...,K\}$ for the primary AI model or the representative number $i\in\{1,...,k\}$ for the XAI model\\
\hline
$f_i$& $i^{th}$ feature\\
\hline
$\mathcal{F}$& The feature set: $\{f_{1}, ..., f_{K}\}$\\
\hline
$F_i^c$& Random variable of $i^{th}$ factor for class $c$, $F_i^c \in \mathbb{R}^{N_c\times 1}$\\
\hline
$F_i$& The $i^{th}$ factor; vertical concatenation of factors $\{F_i^1, \cdots, F_i^C \}$\\
\hline
$R_i^c$& Random variable of $i^{th}$ representative for class $c$, $R_i^c\in \mathbb{R}^{N_c\times 1}$\\
\hline
$R_i$& The $i^{th}$ representative; vertical concatenation of $\{R_i^1, \cdots, R_i^C \}$\\
\hline
$r^c_{j,i}$& $R_i^c$ value in the $j$th observation, $j\in\{1,...,N_c\}$\\
\hline
$\boldsymbol{r_{i}^c}$& Vector of observations belonging to the random variable $R^c_i$\\
\hline
$\boldsymbol{R^c}$ & $N_c\times k$ observation matrix consisting of $r_{j,i}^c$ values arranged so that the column $i$ consists of $\boldsymbol{r_{i}^c}$'s values\\
\hline
$M_{i}^c$&Number of modes of $R^c_i$ in its MMG distribution \\
\hline
$w_{i}$&Importance coefficient of $R_i$\\
\hline
$\lambda_i^c$& Eigenvalue of $F^c_i$\\
\hline
\end{tabular}
\label{symbols}
\end{table}

\subsection{Problem Formalization} 
To formalize the problem that the primary AI is solving, TRUST XAI produces a set of vectors called ``representatives” from the features of the training set. We make three assumptions about the dataset and the representatives as follows.

\textit{\textbf{Assumption 1:}} The statistical inferences made by the XAI model must represent the data’s characteristics. Therefore, the data used to train the primary AI is also used to build the TRUST model. If the primary AI model has to be retrained due to a significant change in the data’s characteristics, the core of the explainer must be rebuilt as well.

\textit{\textbf{Assumption 2:}} The representatives are mutually independent. This allows their joint probability function to be the multiplication of individual probability functions. This assumption simplifies our XAI technique and makes it easier for human users to understand. We select representatives so that the correlation among them is close to zero.

\textit{\textbf{Assumption 3:}} 
Suppose for the $i^{th}$ representative $R_i$, we have a sample $\boldsymbol{r_{i}^c}$ with $N_c$ observations belonging to the class $c$. The distribution of these observations can be approximated with reasonable accuracy by a one-dimensional multi-modal Gaussian (MMG) distribution with $M_{i}^c$ sub-populations, which are called \textit{modes}, with $\{\mu_{i,1}^c, ..., \mu_{i, M^c_i}^c \}$ means and $\{\sigma_{i,1}^c, ..., \sigma_{i, M^c_i}^c\}$ standard deviations. In the case of a unimodal distribution, $M_{i}^c$ is equal to one. 

\subsection{Determining the Representatives} \label{representatives} 

Since factor analysis \cite{Pages2015} satisfies the requirements mentioned in Assumption 2 and more, we use it to determine the representatives. It provides a linear combination of the dataset's features. It also reduces the redundancy in the feature space, which makes our model less likely to overfit. Further, since these new independent latent variables are combinations of the features, we still preserve the most valuable parts of all features. Factor analysis projects the feature values onto axes that maximize the percentage of explained variance in the data. 
Another advantage of this technique is that the resulting factors are orthogonal. Therefore, they are independent with zero covariations.

More specifically, we use factor analysis of mixed data (FAMD) that takes into account different types of variables (quantitative and qualitative) in a dataset \cite{Pages2015}. FAMD is considered the general form of factor analysis. Generally speaking, the core of FAMD is based on principal component analysis (PCA) when variables are all quantitative and multiple correspondence analysis (MCA) when all the variables are qualitative. 

When dealing with a mixed dataset, the typical approach in the research community is either neglecting or encoding (e.g., one-hot encoding) the categorical data when using PCA or transforming the quantitative variables into qualitative variables by breaking down their variation intervals into different levels when using MCA. Despite being relatively easy to implement, these approaches are not accurate enough, and we lose lots of information from the data. Therefore, FAMD is considered the most accurate form of analysis when dealing with a mixed dataset because it takes into account both types of data \cite{Kassambara2017}.

Suppose we have a feature set $\mathcal{F} = \{f_{1}, ..., f_{K}\}$ of mixed data, where $K$ is the total number of the features. After applying factor analysis to the observations belonging to each class separately, we have $K$ factors determined for each class. Note that the number of features and factors are both $K$. For example, for class $c$, factors $\{F_{1}^c, ..., F_{K}^c\}$ are produced.  
The pseudo-code of this function, \texttt{factorAnalysis}, is presented as Algorithm \ref{FA_ALG}. In this algorithm, each data instance belongs to one of the $C$ classes.

In Algorithm \ref{FA_ALG}, $\rho_{f_i, f_j}^2$ is the Pearson correlation coefficient, $\chi_{f_i, f_j}^2$ is the chi-square, and $\eta_{f_i, f_j}^2$ is the squared correlation ratio between feature $i$ and feature $j$ from the standardized matrix of $X^c$ \cite{Jain1991, Maxwell2018}. When measuring the relationship between two quantitative variables Pearson correlation coefficient is used, and for two qualitative variables chi-square is used. Further, when dealing with one variable from each kind, we use the general form of the squared correlation ratio, which results in the Pearson correlation coefficient if both variables are quantitative or chi-square if they are both qualitative. Simply put, these three methods all boil down to the same result, measuring correlations between each pair of variables building a relation matrix for each class. Lastly, singular value decomposition (SVD) is applied to the relation matrix ${RM}^c$ for factorization, removing the interdependencies among the features and projecting them in orthogonal dimensions.

\begin{algorithm}
\caption{Factor Analysis} \label{FA_ALG}
\fontsize{8}{10}\selectfont
\begin{algorithmic}[1]
\REQUIRE  \texttt{factorAnalysis}
\ENSURE Training Set: $X\neq \emptyset \in \mathbb{R}^{N\times K}$; \\AI Model: $\mathcal{AI}$;  Features Set: $\mathcal{F}$
\STATE $y \leftarrow \mathcal{AI}(X)$
\STATE Set labels of each row in $X \leftarrow y$
\STATE Divide $X$ into $X^c \in \mathbb{R}^{N_c\times K}$ sets per class
\FOR {each $X^c$}
\STATE ${X^c}^\prime \leftarrow $ standardized ($X^c$)  \COMMENT{ So it has zero mean and unit variance} 
\STATE Build the relationship matrix (${RM}^c$) such that:
\FOR {each pair $(i,j)$ in $\mathcal{F}(X^c)$} 
\IF{$f_i$ \& $f_j$ are both quantitative}
\STATE {${RM}^c_{i,j} = \rho_{f_i, f_j}^2$}
\ELSIF{$f_i$ \& $f_j$ are both qualitative}
\STATE{${RM}^c_{i,j} = \chi_{f_i, f_j}^2$}
\ELSE
\STATE{${RM}^c_{i,j} = \eta_{f_i, f_j}^2$}
\ENDIF
\ENDFOR
\STATE{$\{{F_{1}^c}^\prime, ..., {F_{K}^c}^\prime\}$ $\&$ $\{\lambda_1^c, ..., \lambda_K^c\} \leftarrow$ SVD(${RM}^c$)}
\STATE $\{F_{1}^c, ..., F_{K}^c\} \leftarrow $ unstandardized ($\{{F_{1}^c}^\prime, ..., {F_{K}^c}^\prime\}$)
\RETURN $\{F_{1}^c, ..., F_{K}^c\}$ 
\ENDFOR
\end{algorithmic}
\end{algorithm}

Then, in Algorithm \ref{PR_ALG}, we use the factors calculated from Algorithm \ref{FA_ALG} to pick the ``representatives" based on how important each factor is.

Factor analysis (Algorithm \ref{FA_ALG}) produces an eigenvalue $\lambda_i^c$ for each factor $F_{i}^c$, which is equivalent to the variation in the data that the factor explains. However, the effectiveness of a factor in distinguishing the class label is not related to the percentage of the explained variation of the data by that factor. Therefore, the percentages of explained variation cannot be used as the importance coefficients for the factors. We elaborate more on this in the following.

\begin{figure}
\begin{center}
  \includegraphics[width=.6\linewidth]{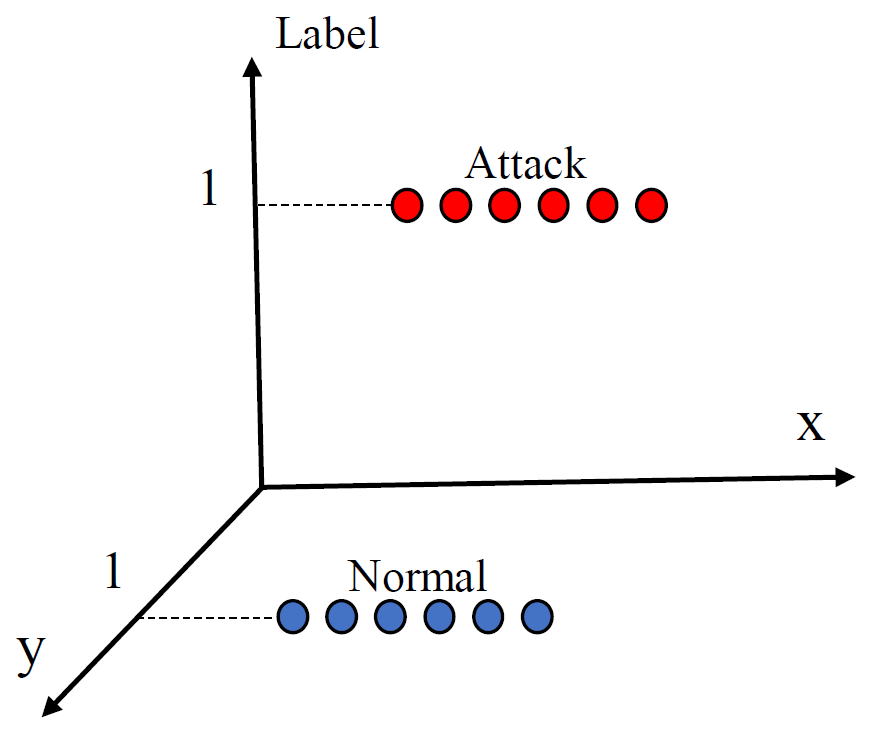}
  \caption{Two-dimensional data with features highly correlated with the labels}
  \label{correlation}
\end{center}
\end{figure}

Suppose we have two-dimensional data with their class labels on the third axis. An example is demonstrated in Fig. \ref{correlation}. The attack class is shown by the red points and the normal class by the blue points. As seen in this figure, since the variation in the $y$-axis values per class is not large, the $y$-axis feature would not be considered important. Therefore, in the case of a dimension reduction or factor analysis, separately for each class, the $x$-axis values would be the primary axis representing each class’s samples. This is because $x$-axis explains almost all the variations in the data points for each class. Meanwhile, it is trivial to see that their $y$-axis values could easily distinguish the class labels. Hence, removing them or treating them as not important will cause a significant loss in the accuracy of the proposed model.

Therefore, we devised a method using the correlation between each factor and the class label to rank the factors and pick the ones with the highest correlation. We use mutual information (MI), which quantifies the amount of information we can get about one variable by knowing the other. It measures the mutual dependence between two variables, which in our case are the factors and the class labels. MI between the factors and the class labels is calculated as

\begin{equation} \label{MI}
\resizebox{.26 \textwidth}{!}
{$\text{MI}(y, F_{i}) = \text{H}(y) - \text{H}(y|F_{i})$}
\end{equation}
where $F_{i} = {[{F_{i}^1}^T \hdots {F_{i}^C}^T]} ^T$ is the vertical concatenation of the $i$th factor values for all classes. The $T$ superscript denotes a transpose of the vector/matrix. $\text{MI}(y,F_{i})$ is the mutual information between the class label vectors $y$ and $F_{i}$. $\text{H}(.)$ is the entropy function, which is defined as follows.

\begin{equation} \label{Entropy}
\resizebox{.33 \textwidth}{!}
{$\text{H}(y) = - \sum_c  P(y = c) \log(P(y = c))$}
\end{equation}

For the sake of simplicity and not having to use differential entropy (since $F_i$ is not discrete), we bin the values in $F_i$, and  $\zeta_{i}$'s are the possible outcomes of the binned random variable $F_{i}$. Suppose $P(\zeta_{i})$ is the probability of $F_{i} = \zeta_{i}$. The conditional entropy $\text{H}(y|F_{i})$ is calculated as follows.

\begin{equation} \label{conditionentropy}
\begin{split}
&\text{H}(y|F_{i}) = \\
& \sum_{\zeta_{i}} P(F_{i} = \zeta_{i}) \text{H}(y|F_{i} =  \zeta_{i}) =  \\
& - \sum_{\zeta_{i}} P(F_{i} = \zeta_{i}) \sum_{c} P(y = c| F_{i} =  \zeta_{i}) \log P(y = c| F_{i} =  \zeta_{i}) \\
 & =  - \sum_{\zeta_{i}} \sum_{c} P(c , \zeta_{i}) \log \frac{P(c,\zeta_{i})}{P (\zeta_{i})}
\end{split}
\end{equation}

We use the MI values as the importance coefficients for the factors. For instance, $w_i$, the importance coefficient of $F_i$, is equal to $\text{MI}(y,F_{i})$. For more details about MI, we refer to \cite{Cover2006}.

Then, we pick only the first $k$ factors, $k\ll K$, with the highest $w_i$ values as the ``representatives." This means the representatives are the factors that have the highest correlation with the class labels. They represent each class's feature values and behavior. 

There is no need to represent each class by its all $K$ factors. Based on the application constraints, we can choose $k$. Regarding the choice of $k$ (i.e., the number of representatives), there is a trade-off between time consumption and performance. For real-time applications, we can pick a lower $k$ (meaning a few factors with the highest importance coefficients as the representatives). For applications where high accuracy is important, we can go with a higher $k$. Since the density and mode computations take time, the more representatives we have, the slower the model would be, but at the same time, it leads to higher accuracy. The time consumption analysis is provided later in this paper. The algorithm to select the representatives is summarized in Algorithm \ref{PR_ALG}.

\begin{algorithm}
\caption{Picking Representatives} \label{PR_ALG}
\fontsize{8}{10}\selectfont
\begin{algorithmic}[1]
\REQUIRE  \texttt{pickingReps} 
\ENSURE Training Set: $X\neq \emptyset \in \mathbb{R}^{N\times K}$; \\AI Model: $\mathcal{AI}$;  Features Set: $\mathcal{F}$
\STATE $\{F_{1}^c, ..., F_{K}^c\} \leftarrow \texttt{factorAnalysis}(X, \mathcal{AI}, \mathcal{F})$
\STATE $y \leftarrow \mathcal{AI}(X)$
\FOR {each $i$}
\STATE $F_{i} = {[ {F_{i}^1}^T, \cdots, {F_{i}^C}^T]}^T$
\STATE $ {w}_i \leftarrow  \text{MI}(y,F_{i})$ 
\STATE ${{w}_i}^\prime \leftarrow$ sort ${w}_i$ in descending order
\STATE $\{R_{1}^c, ..., R_{k}^c\} \leftarrow $  Pick top $k$ factors with the highest ${w_i}^\prime$s
\RETURN $\{R_{1}^c, ..., R_{k}^c\}$ 
\ENDFOR
\end{algorithmic}
\end{algorithm}

\subsection{Density Estimation} 
To study the statistical attributes of the representatives' values, we approximate their density functions with one-dimensional MMG distributions.

Suppose for all the data instances of the training set labeled as the class $c$ by the primary AI model, $X^c$, representative $R_{i}^c$ $i=1,.., k$ has $M_i^c$ modes. For now, assume we know the value of $M_i^c$; we will discuss how to calculate it later in the next section. The one-dimensional multi-modal probability density functions (pdf) of $R_{i}^c$ can be modeled as follows.

\begin{equation} \label{density_function}
p_{i}(R_i^c)=\sum_{m=1}^{M_i^c} \gamma^c_{i,m}  \, \mathcal{N}(R_i^c | \mu_{i,m}^c, \sigma_{i,m}^c)
\end{equation}

$\gamma^c_{i,m} $ is the component weight of sub-population $m$ for $R_{i}^c$ observations in class $c$. $\mathcal{N}$ is a normal distribution with mean $\mu_{i,m}^c$ and standard deviation $\sigma_{i,m}^c$. To make the total area under the pdf normalized for every representative per class, for $\forall c \vee \forall i$, $\sum_{m}\gamma^c_{i,m} = 1$ should be satisfied. Here we use the expectation-maximization (EM) algorithm \cite{ng2012algorithm} to fit the representatives' values to a proper distribution in the form of Eq. \ref{density_function}.

\begin{figure}[t]
\centering
  \includegraphics[scale=0.08]{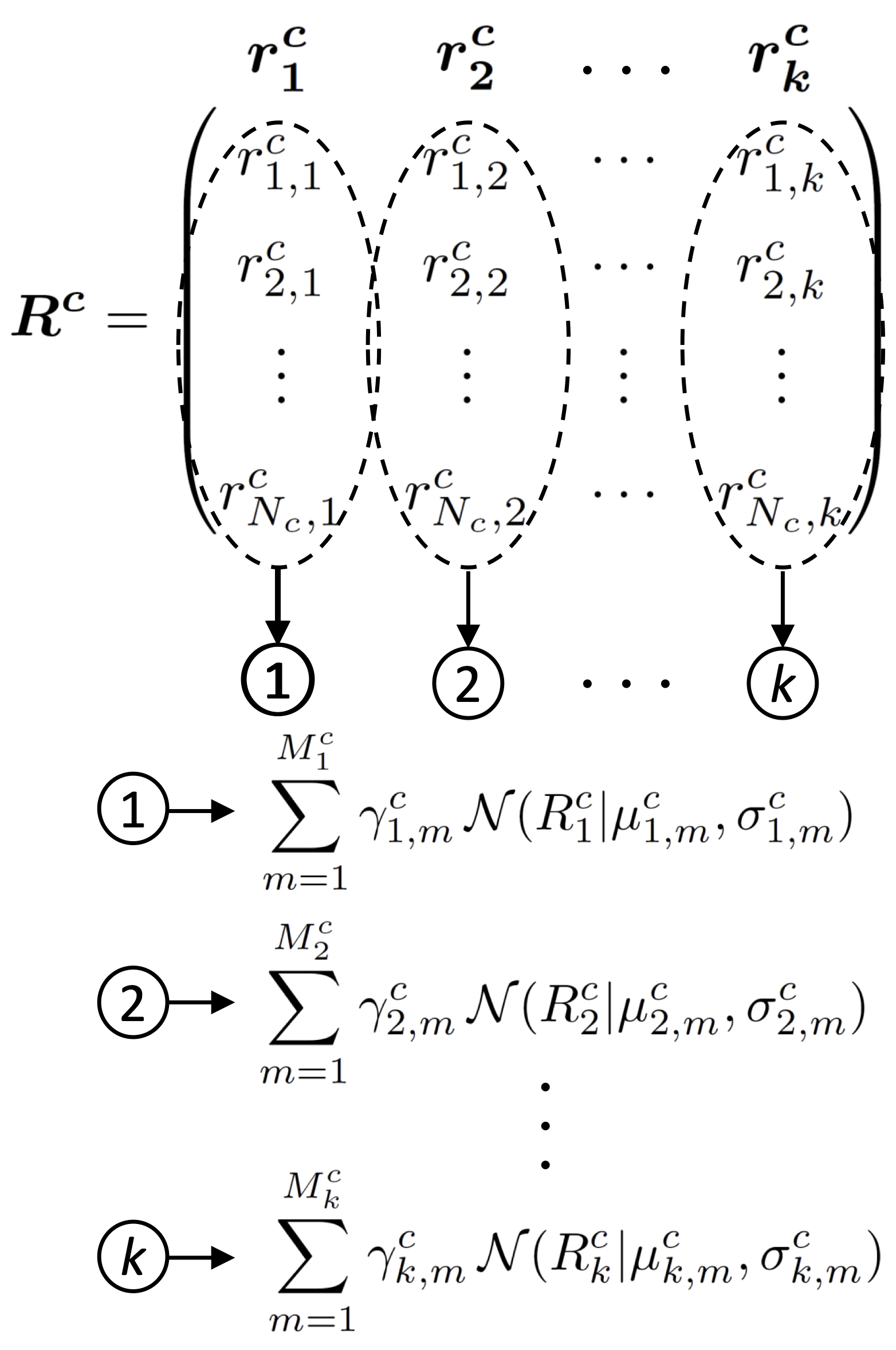}
  \caption{Approximating representatives to unique MMG distributions for class $c$}
  \label{multi-modal}
\end{figure}

After the above process, we end up with $k \times C$ distributions. These distributions build the backbone of the TRUST explainer.  Figure \ref{multi-modal} shows the approximation process for the values of class $c$ representatives as an example. The $\boldsymbol{R^c} = [\boldsymbol{r_{1}^c}, \boldsymbol{r_{2}^c}, ..., \boldsymbol{r_{k}^c}]\in \mathbb{R}^{N_c\times k}$ equates to the matrix of observations of the representatives of the class $c$, where $R_i^c$ is the  random variable for the set of observations $\boldsymbol{r_{i}^c}$.

Now, it is time to show how our TRUST XAI explains the data instances labeled as a specific class by the primary AI model. First, the likelihoods that the new instance's representatives belong to each class's representatives are calculated. These likelihoods can be estimated from the pdf models. The total likelihood is the product of the likelihoods of all representatives. We apply a logarithm transformation to decrease the computational complexity and compute the log-likelihoods. Thus, we simplify the computations from multiplications to additions (which is less expensive), avoid computation of the exponentials, and prevent any arithmetic underflow or overflow (since likelihoods could become very large or small). Log transformation also reduces or removes the skewness in the data. Suppose the new instance is $\boldsymbol{z} = [z_1, ..., z_K]\in \mathbb{R}^ {1 \times K}$. The inputs to TRUST XAI are only the representatives' values. Therefore, $\boldsymbol{z}$ is projected on the factors' space and cropped to its corresponding representatives' values $\boldsymbol{z^\prime} = [z_1^\prime, ..., z_k^\prime]\in \mathbb{R}^ {1 \times k}$. The conditional likelihood of $z_i^\prime$, $i=1, .., k$ belonging to class $c$ can be written as:

\begin{equation}
\begin{aligned}
&p_i(z_i^\prime|c)= \\
&\! \!\sum_{m=1}^{M^c_i}\!\exp\!\bigg(\!\log(\gamma^c_{i,m})\!+\!\log(\mathcal{N}(z_i^\prime | \mu_{i,m}^c, \sigma_{i,m}^c))\!\!\bigg)
\end{aligned}
\label{re-written}
\end{equation}

After taking a logarithm from Eq. \ref{re-written}, the conditional log-likelihood of $z_i^\prime$ belonging to $R_i^c$ is calculated by Eq. \ref{log_likelihood}.

\begin{equation}
\begin{aligned}
&\log (p(z_i^\prime|c))= \\
&\log\! \bigg(\!\sum_{m=1}^{M^c_i}\!\exp\!\Big(\!\log(\gamma^c_{i,m})\!+\!\log(\mathcal{N}(z_i^\prime | \mu_{i,m}^c, \sigma_{i,m}^c))\!\Big)\!\!\bigg)=\\
&\log\! \bigg(\!\sum_{m=1}^{M^c_i}\!\exp\!\Big( \alpha^c_{i,m} - \frac{1}{2} (\frac{z_i^\prime - \mu_{i,m}^c}{\sigma_{i,m}^c})^2\!\Big)\!\!\bigg)
\end{aligned}
\label{log_likelihood}
\end{equation}

where,

\begin{equation}
\alpha^c_{i,m} = \log(\gamma^c_{i,m}) - \log(\sigma_{i,m}^c)- \frac{1}{2} \log(2\pi)
\end{equation}

$\alpha^c_{i,m}, \forall c \vee \forall i \vee \forall m$ is a fixed constant.

Subsequently, we calculate the weighted sum of these log-likelihoods using the representatives' importance coefficients (the MI values, ${w_i}$, from Subsection B). The ${w_i}$'s are normalized before the calculation, though. The importance coefficient of the representative $R_i$ is calculated by:

\begin{equation}
\hat{w_i} = \frac{w_i}{\sum_{j = 1}^{k}w_j}
\end{equation}

Since the $w_i$'s are normalized, $\sum_{i = 1}^{k}\hat{w_i} = 1$. After that, the log-likelihood of classifying $\boldsymbol{z}^\prime$ as class $c$ is computed as:

\begin{equation} \label{log(p|c)}
\log(p(\boldsymbol{z}^\prime|c)) = \sum_{i= 1}^{k} \hat{w_i} \times \log (p_i(z_i^\prime|c))
\end{equation}

Finally, the predicted class for $z$ by the primary AI, represented as $\ell_{z}$, is explained by TRUST XAI as the class with the maximum total log-likelihood:

\begin{equation} \label{maxlog}
\ell_{\boldsymbol{z}} = \max_{c} \log(p(\boldsymbol{z}^\prime|c))
\end{equation}

It is essential to notice that we are not following the common clustering processes that model the whole training set as a multi-dimensional (with features as the dimensions) multi-cluster distribution and then calculate the likelihood of the new data belonging to each cluster (i.e., class). Here, we take a completely different approach. We approximate the one-dimensional distribution of each representative and then calculate the likelihood of the new instance's representative values belonging to the corresponding representative distributions for each class. Afterward, by the weighted sum of these likelihoods for each class, we have $C$ likelihoods, each resulting from $k$ distributions. Then, the class with the highest likelihood is picked as the explained class.

\subsection{Selecting the Number of Modes} \label{Mode_Selection}
A critical parameter in the TRUST explainer is the number of modes for MMG distribution (i.e., $M^c_i$) for each representative (as mentioned in Assumption 3). In general, this number is different for each representative, and it can be optimized to increase speed and performance. We use the grid search technique which is a standard method for determining hyperparameters in machine learning techniques. Through grid search, for each representative, we find the number of modes that maximizes the probability of the correct prediction for samples from different classes. The choice of  $M^c_i$ for each representative is important since it has a critical role in how well the fitted MMG distribution works as a density estimator. The pseudo-code of the method is shown in Algorithm \ref{GS_ALG}. This algorithm searches over a zone called $Z$, which is simply a grid of integers starting from 1 to any potential number of modes.

\begin{algorithm}[t]
\caption{Grid Mode Selection} \label{GS_ALG}
\fontsize{8}{10}\selectfont
\begin{algorithmic}[1]
\REQUIRE \texttt{modeSelect} 
\ENSURE Representatives of all the classes:
\\ $\{R_{1}^1, ..., R_{k}^1\} \in \mathbb{R}^{N_1\times k},  \cdots, \{R_{1}^C, ..., R_{k}^C\} \in \mathbb{R}^{N_C\times k}$; 
\\ Zone $Z \in \mathbb{R}^{C \times C}$ 
\STATE{$y \leftarrow {[{\text{ones}(N_1)}^T, \hdots, C \times{\text{ones}(N_C)}^T]}^T$}
\FOR {each set of $(R_{i}^1 \cdots R_{i}^C)$}
\STATE {$\mathit{Max} \leftarrow 0$}
\FOR {$ m^1$ in $Z$} 
\STATE {$\text{MMG}^1_i  \leftarrow \text{fit } R_{i}^1 \text{ to a Gaussian with } m^1 \text{ modes} $}
\STATE$\vdots$
\FOR {$m^C$ in $Z$}
\STATE {$\text{MMG}^C_i  \leftarrow \text{fit } R_{i}^C \text{ to a Gaussian with } m^C \text{ modes} $}
\STATE{$P_{m^1} \leftarrow p_1(R_i|\text{MMG}^1_i )$}   
\STATE$\vdots$	\hfill\COMMENT{$\tiny {R_i = {[{R_{i}^1}^T, \hdots, {R_i^C}^T ]}^T}$}
\STATE{$P_{m^C} \leftarrow p_C(R_i|\text{MMG}^C_i )$}
\STATE{$P\leftarrow [P_{m^1} \cdots P_{m^C} ]$}
\STATE{$y_\text{MMG}  \leftarrow  \text{index}(P.\max(1))$}
\STATE{$ \text{score}_{m^1, \cdots, m^C} \leftarrow \text{MCC}(y_\text{MMG}, y)$} 
\IF{$\text{score}_{m^1, \cdots, m^C} > \mathit{Max}$}
\STATE {$\mathit{Max} \leftarrow \text{score}_{m^1, \cdots, m^C}$}
\STATE{$M^1_i \leftarrow m^1$}
\STATE$\vdots$
\STATE{$M^C_i \leftarrow m^C$}
\ENDIF
\ENDFOR
\STATE$\vdots$
\ENDFOR
\ENDFOR
\RETURN $\{M_{1}^1, ..., M_{k}^1\}$ $\cdots$ $\{M_{1}^C, ..., M_{k}^C\}$
\end{algorithmic}
\end{algorithm}

As seen in Algorithm \ref{GS_ALG}, the dimension of the search space grows exponentially with the number of classes. Also, if the number of modes or classes is large, the search zone would be too large. Hence, the search could be very slow and time-consuming. We have developed a faster algorithm that divides the search zone into smaller sub-zones, as is shown in Algorithm \ref{FGS}. At first, the probabilities of the centers of each sub-zone as the number of modes are calculated. Then the sub-zone with the highest score is chosen to be searched thoroughly. Even though this algorithm might not give us the exact number of modes, it provides a good accuracy (in the order of 99\% in our experiments). With decreasing the size of sub-zones, the accuracy increases. 

As seen in line 14 of Algorithm \ref{GS_ALG} and line 15 of Algorithm \ref{FGS}, Matthew's correlation coefficient (MCC), Eq. \ref{mcc}, is used to calculate the score. We have not used the Accuracy metric, Eq. \ref{accu}, because MCC has been shown to be a better metric when dealing with imbalanced datasets, which are common in cybersecurity and other numerical applications \cite{Zolanvari2018}. The undetected rate (UR), Eq. \ref{ur}, is another useful metric for performance evaluation used in the next section. 

\begin{equation} \label{mcc}
\resizebox{.4 \textwidth}{!}
{$\text{MCC} = \frac{TP \times TN - FP \times FN}{\sqrt{(TP+FP)(TP+FN)(TN+FP)(TN+FN)}}$}
\end{equation}

\begin{equation} \label{accu}
\resizebox{.25 \textwidth}{!}
{$\text{Accuracy} = \frac{TP+TN}{TP+TN+FP+FN}$}
\end{equation}

\begin{equation} \label{ur}
\resizebox{.12 \textwidth}{!}
{$\text{UR} = \frac{FN}{FN+TP}$}
\end{equation}

\noindent where $TN$ is the number of normal data labeled as normal, $TP$ is the number of attack data classified as attack, $FP$ is the number of normal labeled as attack, and $FN$ is the number of attacks labeled as normal by the model.

\begin{algorithm} [t]
\caption{Fast Grid Search} \label{FGS}
\fontsize{8}{10}\selectfont
\begin{algorithmic}[1]
\REQUIRE  $\texttt{modeSelectFast}$
\ENSURE Representatives of all the classes:
\\ $\{R_{1}^1, ..., R_{k}^1\} \in \mathbb{R}^{N_1\times k},  \cdots, \{R_{1}^C, ..., R_{k}^C\} \in \mathbb{R}^{N_C\times k}$; 
\\ Zone $Z \in \mathbb{R}^{C \times C}$ 
\STATE{$y \leftarrow {[{\text{ones}(N_1)}^T, \hdots, C \times{\text{ones}(N_C)}^T]}^T$}
\FOR {each set of $(R_{i}^1 \cdots R_{i}^C)$}
\STATE {Divide zone $Z$ into $z_1, ..., z_d$}
\STATE {$\mathit{Max} \leftarrow 0$} 
\FOR {each $z_j$}
\STATE {$(m^1, \cdots,m^C) \leftarrow z_j \text{'s center}$}
\STATE {$\text{MMG}^1_i  \leftarrow \text{fit } R_{i}^1 \text{ to a Gaussian with } m^1 \text{ modes} $}
\STATE$\vdots$
\STATE {$\text{MMG}^C_i  \leftarrow \text{fit } R_{i}^C \text{ to a Gaussian with } m^C \text{ modes} $}
\STATE{$P_{m^1} \leftarrow p_1(R_i|\text{MMG}^1_i )$}   
\STATE$\vdots$	\hfill\COMMENT{$\tiny {R_i = {[{R_{i}^1}^T, \hdots, {R_i^C}^T ]}^T}$}
\STATE{$P_{m^C} \leftarrow p_C(R_i|\text{MMG}^C_i )$}
\STATE{$P\leftarrow [P_{m^1} \cdots P_{m^C} ]$}
\STATE{$y_\text{MMG}  \leftarrow  \text{index}(P.\max(1))$}
\STATE{$ \text{score}_{m^1, \cdots, m^C} \leftarrow \text{MCC}(y_\text{MMG}, y)$}
\IF{$\text{score}_{m^1, \cdots, m^C} > \mathit{Max}$}
\STATE {$\mathit{Max} \leftarrow \text{score}_{m^1, \cdots, m^C}$}
\STATE{$Z_i \leftarrow z_j$}
\ENDIF
\ENDFOR
\STATE $\texttt{modeSelect}(\{ R_{i}^1\}, \cdots, \{R_{i}^C\}, Z_i)$ 
\ENDFOR
\RETURN $\{M_{1}^1, ..., M_{k}^1\}$ $\cdots$ $\{M_{1}^C, ..., M_{k}^C\}$
\end{algorithmic}
\end{algorithm}

\section{Case Study: IIoT Security }
Sophisticated AI models have been vastly applied in intrusion detection systems (IDS). XAI would add interpretability and trust to these systems. We have built a lab-scaled IIoT system to collect realistic and up-to-date datasets for network-based IDSs. We have implemented a supervisory control and data acquisition (SCADA) system widely used by industries to supervise the level and turbidity of liquid storage tanks. This IIoT system is employed in industrial reservoirs and water distribution systems as a part of the water treatment and distribution. For more information regarding our testbed, we refer the readers to our previous papers \cite{Zolanvari2019} and \cite{Teixeira2018}.

\subsection{Utilized Datasets}
The first tested dataset is the one collected from our testbed, which we refer to as ``WUSTL-IIoT."  To collect the proper dataset, we (as a white-hat attacker) attacked our testbed using manipulated commands, such as backdoor, command injection, denial of service, and reconnaissance \cite{Zolanvari2019}. 

Further, we have tested our TRUST XAI on two other cybersecurity datasets to add an empirical proof of concept, ``NSL-KDD" \cite{Tavallaee2009} and ``UNSW" \cite{Moustafa2015}. Specifications of these datasets are shown in Table \ref{three-DS}. NSL-KDD is a revised and cleansed version of the KDD’99 published by the Canadian Institute for Cybersecurity from the University of New Brunswick. KDD'99 was initially collected by MIT Lincoln Laboratory through a competition to collect traffic records and build a network intrusion detector. This dataset consists of four different types of attacks. Here, we label them all as class 1 for binary classification. Further, the UNSW-NB15 dataset, which we refer to as UNSW, was built by the Cyber Range Lab of the Australian Centre for Cyber Security (ACCS) in 2015. The traffics are labeled as either 0 or 1 for normal or attack class, respectively.

\begin{table}[t]
\centering\fontsize{9}{11}\selectfont
\caption{Specifics of the three datasets}
\begin{tabular}{|c|c|c|c|}
\hline
\textbf{Dataset} & WUSTL-IIoT & NSL-KDD  &   UNSW \\
\hline
\textbf{\# of observations} & 1,194,464 & 125,973 & 65,535\\
\hline
\textbf{\# of features} & 41 & 40  & 41\\
\hline
\textbf{\# of attacks} & 87,016 & 58,630 & 45,015 \\
\hline
\textbf{\# of normals} &1,107,448 & 67,343 & 20,520 \\
\hline
\end{tabular}
\label{three-DS}
\end{table}

\subsection{The Primary AI Model}
TRUST XAI is a model-agnostic technique that can be applied to any complex AI model. To evaluate its performance on IIoT security, we use an artificial neural network (ANN) as a binary classifier with two hidden layers with 20 and 10 neurons respectively with \textit{ReLU} activation functions, ten epochs, with \textit{adam} optimizer, and \textit{sigmoid} activation function in the output layer. The scikit-learn library \cite{Pedregosa2011} was used to implement the ANN model (as the primary AI model). ANN is generally considered complex and unexplainable to human users. The inputs to the IDS are the flow instances as mentioned in the previous subsection. The output of the IDS can be a multi-class or binary classification. However, for the sake of simplicity, we have built the TRUST explainer by treating the outputs of the IDS as binary classes with normal as 0 and attack as 1. 

Even though the performance of the primary AI is not of our interest, we show its results on the three datasets in Tables \ref{accuracy_AI_wustl}, \ref{accuracy_AI_nsl}, and \ref{accuracy_AI_unsw} for comparison. Each dataset is split into training and test sets with an 80:20 ratio. The performance on the training set is presented in the (a) tables, while the test results are shown in the (b) tables.

\begin{table}[H]  
  \caption{Performance of the primary AI on WUSTL-IIoT}
  \begin{subtable}{.5\linewidth}
  \centering\fontsize{9}{11}\selectfont
  \begin{tabular}{|c|c|c|}
    \cline{2-3}
    \multicolumn{1}{c|}{} & Normal & Attack \\ \hline
    Normal& 885980  & 12   \\ \hline
    Attack & 131    & 69448    \\ \hline
  \end{tabular}
  \caption{On the training set}
   \label{accuracy_AI_Training}
   \end{subtable}%
    \begin{subtable}{.5\linewidth}
      \centering\fontsize{9}{11}\selectfont
    \begin{tabular}{|c|c|c|}
    \cline{2-3}
    \multicolumn{1}{c|}{} & Normal & Attack \\ \hline
   Normal& 221452  & 4   \\ \hline
    Attack & 40    & 17397    \\ \hline
  \end{tabular}
   \caption{On the test set}
   \label{accuracy_AI_Testing}
   \end{subtable}
   \label{accuracy_AI_wustl}
  \end{table}

\begin{table}[H]  
  \caption{Performance of the primary AI on NSL-KDD}
  \begin{subtable}{.5\linewidth}
  \centering\fontsize{9}{11}\selectfont
  \begin{tabular}{|c|c|c|}
    \cline{2-3}
    \multicolumn{1}{c|}{} & Normal & Attack \\ \hline
    Normal& 53664   & 173  \\ \hline
    Attack & 529   & 46412    \\ \hline
  \end{tabular}
  \caption{On the training set}
   \label{accuracy_AI_Training}
   \end{subtable}%
    \begin{subtable}{.5\linewidth}
      \centering\fontsize{9}{11}\selectfont
    \begin{tabular}{|c|c|c|}
    \cline{2-3}
    \multicolumn{1}{c|}{} & Normal & Attack \\ \hline
    Normal& 13452   & 54   \\ \hline
    Attack & 138   & 11551    \\ \hline
  \end{tabular}
   \caption{On the test set}
   \label{accuracy_AI_Testing}
   \end{subtable}
   \label{accuracy_AI_nsl}
  \end{table}

\begin{table}[H]  
  \caption{Performance of the primary AI on UNSW }
  \begin{subtable}{.5\linewidth}
  \centering\fontsize{9}{11}\selectfont
  \begin{tabular}{|c|c|c|}
    \cline{2-3}
    \multicolumn{1}{c|}{} & Normal & Attack \\ \hline
    Normal& 15791   & 657   \\ \hline
    Attack & 473   & 35507    \\ \hline
  \end{tabular}
  \caption{On the training set}
   \label{accuracy_AI_Training}
   \end{subtable}%
    \begin{subtable}{.5\linewidth}
      \centering\fontsize{9}{11}\selectfont
    \begin{tabular}{|c|c|c|}
    \cline{2-3}
    \multicolumn{1}{c|}{} & Normal & Attack \\ \hline
    Normal& 3895   & 177   \\ \hline
    Attack & 115  & 8920  \\ \hline
  \end{tabular}
   \caption{On the test set}
   \label{accuracy_AI_Testing}
   \end{subtable}
   \label{accuracy_AI_unsw}
  \end{table}

These tables have been brought up to emphasize once again that to develop an XAI in general, only the labels outputted by the primary AI are important while ignoring the true labels of the samples. This is because, in the XAI domain, we care about explaining the primary AI's behavior, not how accurately it has behaved.

The results of these tables based on the metrics including Accuracy, MCC, and undetected rate (UR) using Eq. \ref{accu},  Eq. \ref{mcc}, and Eq. \ref{ur} are summarized in Table \ref{AI_summ}.

\begin{table}[H]  
  \centering\fontsize{9}{11}\selectfont
  \caption{Summary of the primary AI's performance}
  \begin{tabular}{|p{3cm}|p{1.2cm}|p{1cm}|p{.8cm}|}
    \cline{2-4}
    \multicolumn{1}{c|}{} & \textbf{Accuracy}  & \textbf{MCC} & \textbf{UR}\\ \hline
    \mbox{WUSTL-IIoT, Training} & 99.98\%  & 99.88\% & 0.19\% \\ \hline
    WUSTL-IIoT, Test  & 99.98\%  &   99.86\% & 0.23\% \\ \hline
    NSL-KDD, Training  & 99.30\%  & 98.60\%  & 1.13\%\\ \hline
	NSL-KDD, Test  & 99.24\%  &  98.47\% & 1.18\%\\ \hline
	UNSW, Training  & 97.84\%  & 94.98\%   & 1.3\% \\ \hline
	UNSW, Test  & 97.77\%  &  94.78\%  & 1.27\%\\ \hline
  \end{tabular}
   \label{AI_summ}
  \end{table}

Next, we discuss the steps to build the core of TRUST from the output labels from the primary AI. These steps include picking representatives and mode selection.

\subsection{The XAI Model: TRUST} 

The first step is to run the \texttt{pickingReps} function using Algorithm \ref{PR_ALG} on the training set with their predicted labels by the primary AI. We end up with $K$ factors. Afterward, the MI scores between each representative and the class labels have been calculated to determine the top $k$ representatives that have the highest correlation with the class labels and their importance coefficients.

As mentioned before, the choice of $k$ provides a trade-off between the speed and the performance of the explainer. Large $k$ may result in a better performance with a lower speed and vice-versa. As shown in Figure \ref{train-test}, just for the sake of experiment, we try $k$ from $1$ to $8$. Please note that $k$ can be any number between $1$ and $K$. As shown later, we get high MCC scores even with $k$ equal to 4 or 5. Therefore, there is no need to sacrifice the speed or complexity of the model by choosing a large $k$.

The next step is to estimate the number of modes per representative's probability density function. The general case with the $C$ number of classes has been demonstrated in Algorithm \ref{GS_ALG}. In our case, we have only two classes, $C = 2$, therefore we have only two sets of representatives, $\{R_{1}^1, ..., R_{k}^1\}$ and $\{R_{1}^2, ..., R_{k}^2\}$. The modified mode computation algorithms for two classes are shown in Algorithm \ref{GS_ALG2} and Algorithm \ref{FGS2}.

\begin{algorithm}
\caption{Grid Mode Selection with $C=2$} \label{GS_ALG2}
\fontsize{8}{10}\selectfont
\begin{algorithmic}[1]
\REQUIRE  $\texttt{modeSelectC2}$ 
\ENSURE Representatives of class "1": $\{R_{1}^1, ..., R_{k}^1\} \in \mathbb{R}^{N_1\times k}$;
\\Representatives of class "2": $\{R_{1}^2, ..., R_{k}^2\} \in \mathbb{R}^{N_2\times k}$; \\ Zone $Z \in \mathbb{R}^{2 \times 2}$
\FOR {each $(R_{i}^1,R_{i}^2)$}
\STATE {$\mathit{Max} \leftarrow 0$}
\FOR {$ m^1$ in $Z$} 
\STATE {$\text{MMG}^1_i  \leftarrow \text{fit } R_{i}^1 \text{ to a guassian with } m^1 \text{ modes} $}
\FOR {$m^2$ in $Z$} 
\STATE {$\text{MMG}^2_i  \leftarrow \text{fit } R_{i}^2 \text{ to a guassian with } m^2 \text{ modes} $}
\STATE{$P_{m^1} \leftarrow p_1(R_i|\text{MMG}^1_i )$}   \hfill\COMMENT{$R_i = {R_{i}^1 \brack R_i^2}$}
\STATE{$P_{m^2} \leftarrow p_2(R_i|\text{MMG}^2_i )$}
\STATE{$y_\text{MMG}  \leftarrow P_{m^2} > P_{m^1} $}
\STATE{$ \text{score}_{m^1,m^2} \leftarrow \text{MCC}(y_\text{MMG}, {\text{zeros}(N) \brack \text{ones}(N)})$}
\IF{$\text{score}_{m^1,m^2} > \mathit{Max}$}
\STATE {$\mathit{Max} \leftarrow \text{score}_{m^1,m^2}$}
\STATE{$M^1_i \leftarrow m^1$}
\STATE{$M^2_i \leftarrow m^2$}
\ENDIF
\ENDFOR
\ENDFOR
\ENDFOR
\RETURN $\{M_{1}^1, ..., M_{k}^1\}$ $\&$ $\{M_{1}^2, ..., M_{k}^2\}$
\end{algorithmic}
\end{algorithm}

\begin{algorithm}
\caption{Fast Grid Search with $C=2$} \label{FGS2}
\fontsize{8}{10}\selectfont
\begin{algorithmic}[1]
\REQUIRE  \texttt{modeSelectFastC2}
\ENSURE Representatives of class "1": $\{R_{1}^1, ..., R_{k}^1\} \in \mathbb{R}^{N_1\times k}$;
\\Representatives of class "2": $\{R_{1}^2, ..., R_{k}^2\} \in \mathbb{R}^{N_2\times k}$; \\Zone $Z \in \mathbb{R}^{2 \times 2}$
\FOR {each $(R_{i}^1,R_{i}^2)$}
\STATE {Divide zone $Z$ into $z_1, ..., z_d$}
\STATE {$\mathit{Max} \leftarrow 0$} 
\FOR {each $z_j$}
\STATE {$(m^1,m^2) \leftarrow z_j \text{'s center}$}
\STATE {$\text{MMG}^1_i  \leftarrow \text{fit } R_{i}^1 \text{ to a guassian with } m^1 \text{ modes} $}
\STATE {$\text{MMG}^2_i  \leftarrow \text{fit } R_{i}^2 \text{ to a guassian with } m^2 \text{ modes} $}
\STATE{$P_{m^1} \leftarrow p_1(R_i|\text{MMG}^1_i )$}   \hfill\COMMENT{$R_i = {R_{i}^1 \brack R_i^2}$}
\STATE{$P_{m^2} \leftarrow p_2(R_i|\text{MMG}^2_i )$}
\STATE{$y_\text{MMG}  \leftarrow P_{m^2} > P_{m^1} $}
\STATE{$ \text{score}_{m^1,m^2} \leftarrow \text{MCC}(y_\text{MMG},  {\text{zeros}(N) \brack \text{ones}(N)})$}
\IF{$\text{score}_{m^1,m^2} > \mathit{Max}$}
\STATE {$\mathit{Max} \leftarrow \text{score}_{m^1,m^2}$}
\STATE{$Z_i \leftarrow z_j$}
\ENDIF
\ENDFOR
\STATE {$\texttt{modeSelectC2}(\{ R_{i}^1\},\{R_{i}^2\}, Z_i)$}
\ENDFOR
\RETURN $\{M_{1}^1, ..., M_{k}^1\}$ $\&$ $\{M_{1}^2, ..., M_{k}^2\}$
\end{algorithmic}
\end{algorithm}


\subsection{Results} \label{results}
The representatives of TRUST XAI are built using the training set. The test samples' factors are calculated by projecting their feature values in the factor space computed using the training set. Then the same factors chosen as representatives in the training set are chosen for the test samples as well. Afterward, the likelihoods are calculated for the corresponding representatives, and the class with the highest likelihood is explained as the class of that sample.

\begin{figure}[t]
\centering
  \includegraphics[width=0.5\textwidth]{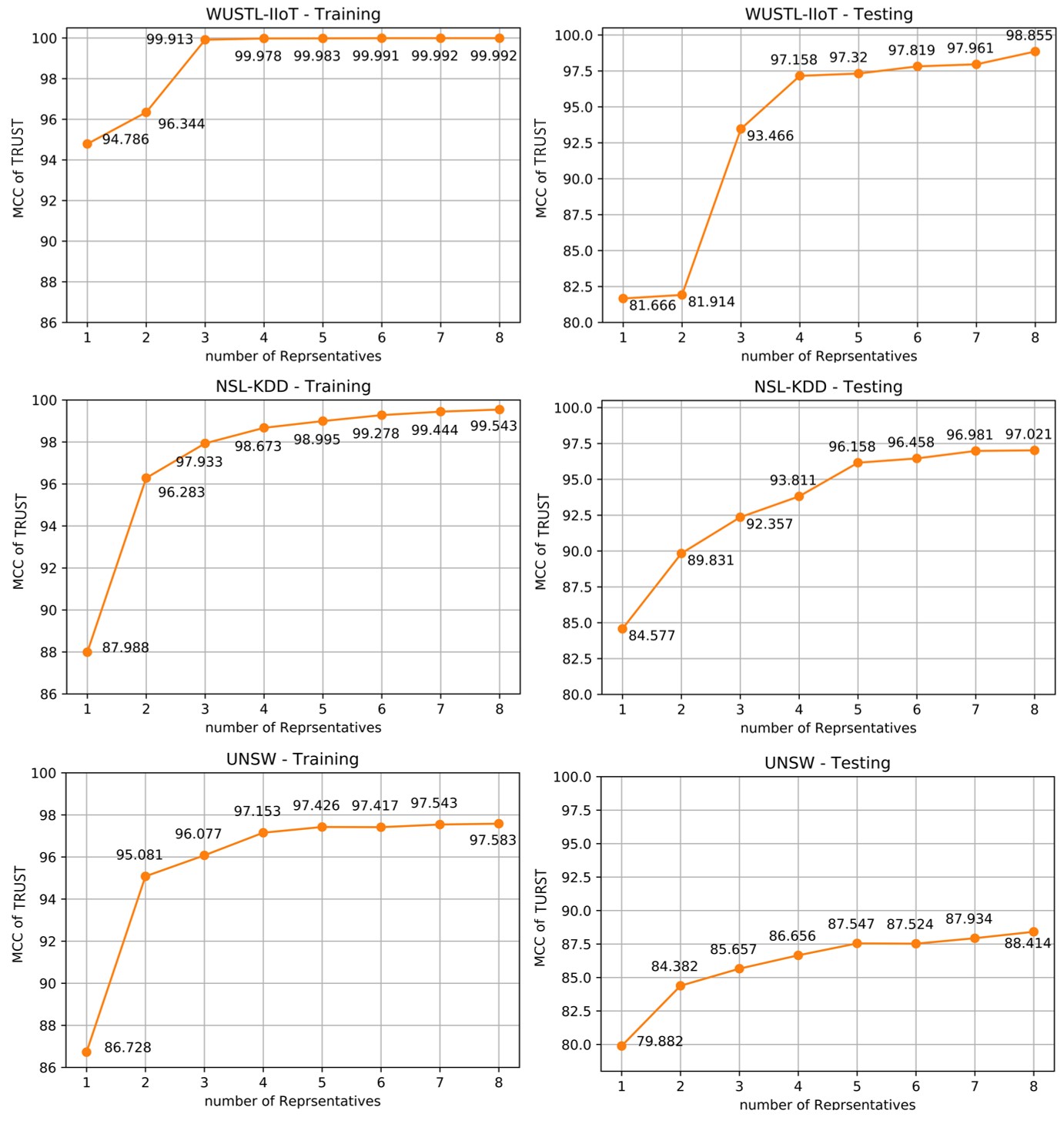}
  \caption{\label{train-test}MCC scores of TRUST XAI on the training sets and the test sets.}
\end{figure}

Figure \ref{train-test} shows the MCC results vs. the number of representatives, $k$, for the training and test sets. Note that MCC values are generally less than accuracy values. The results show TRUST is highly successful in explaining the primary AI's output on unseen data with an average of 90.89\%  MCC or 98\% accuracy.

\begin{figure*}[t]
\centering
  \includegraphics[width=\textwidth]{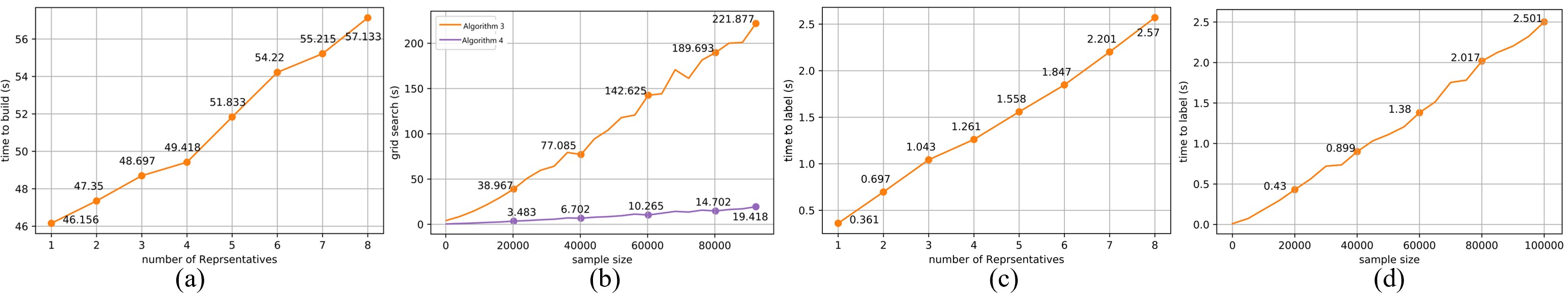}
  \caption{\label{time_con}TRUST's Time consumption of (a) building the core model based on the number of representatives, $k$, (b) comparing Algorithm \ref{GS_ALG} and Algorithm \ref{FGS}, (c) labeling 100,778 samples vs. the number of representatives, $k$, (d) labeling time vs. the number of samples, $N$. In the last two figures, the training set is used because it has a larger number of samples.}
\end{figure*}


We have analyzed the performance in terms of time consumption, using the NSL-KDD dataset as an example. These experiments were run on a standard laptop with an Intel Core i7 CPU, 16 GB RAM, and Windows OS (no GPU). The steps to build the TRUST model only need to be done once, and after that, the model is reusable to easily explain the labels of future observations. Figure \ref{time_con}a shows the computation time to build the TRUST model (including the factor analysis, picking the representatives, and estimating their density functions) versus the number of representatives, $k$, was chosen to build the TRUST model. The time to search for the number of modes is not included in this graph. As expected, by increasing $k$, it takes more time to build the model since we have more representatives that they require to calculate their density functions.

Algorithm \ref{GS_ALG} and Algorithm \ref{FGS} to determine the number of modes for a representative are compared in Figure \ref{time_con}b. The time consumption is calculated versus the number of samples. The search zone is a 20 by 20 grid, and when using Algorithm \ref{FGS}, it was divided into sixteen 5×5 sub-zones. As seen in this figure, Algorithm \ref{FGS} speeds up the mode search by more than ten times.

Next, the time for reasoning the labels of test samples by the TRUST model is examined. This step is done after the core of TRUST is built and is ready to label the unseen data as the class with the highest likelihood. First, TRUST labeling time consumption as a function of different numbers of representatives, $k$, is plotted in Figure \ref{time_con}c. As seen in this figure, the compute time is pretty insignificant. Second, Figure \ref{time_con}d shows the time consumption as a function of the number of samples. 

\begin{figure}[t]
\centering
  \includegraphics[width=0.51\textwidth]{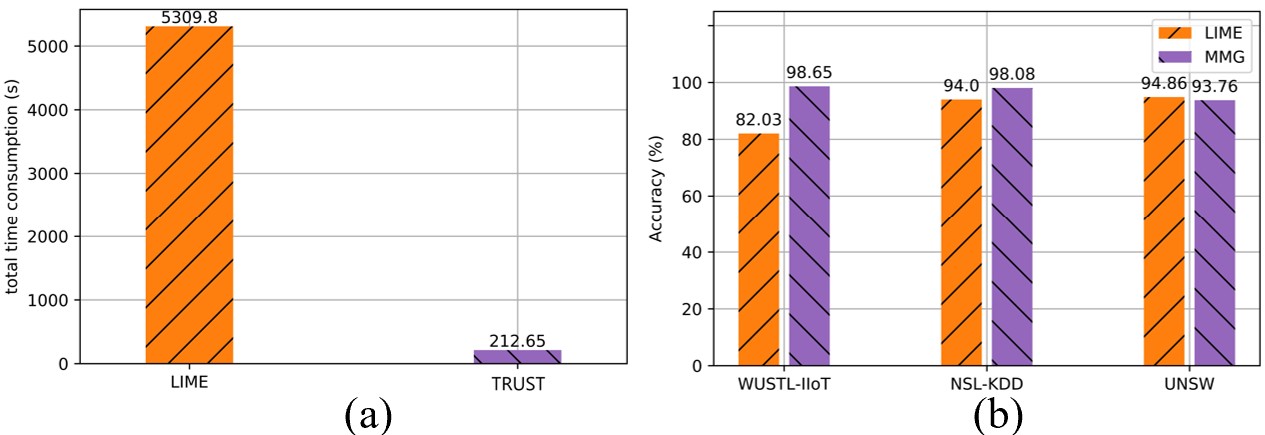}
  \caption{\label{comp_lime_perform}Explaining 8000 test samples; (a) time consumption of our TRUST model vs. LIME, (the TRUST's built time is also included), (b) performance of TRUST vs. LIME.}
\end{figure}

As mentioned before, our model has a significant improvement over the XAI models that require repeating the whole process for every single sample, such as LIME. These repetitions make their model time-consuming and inefficient when dealing with a large number of samples. On the contrary, our technique consists of a core model. Once it is built, it is ready to explain the labels of future unseen data. It is fast and efficient when dealing with a significant amount of unseen incoming data. Figure \ref{comp_lime_perform}a demonstrates this claim. Note that the TRUST’s time of 212.65 s (3.5 minutes) includes the time to build the model from the 100,778 training samples. The time includes factor analysis, picking eight representatives, running the mode grid search (Algorithm \ref{FGS}) for them, estimating their density functions, and labeling the 8000 test samples. Our model is almost 25 times faster than LIME, which makes a significant difference in real-time applications. Next, we have compared the performance of our model with LIME using the accuracy metric, Eq. \ref{accu}. Here, we used only five representatives in our TRUST model and calculated the accuracy of the test labels. As seen from Figure \ref{comp_lime_perform}b, our model has performed better even with only five representatives, except for UNSW, where LIME has slightly higher accuracy (94.86\% vs. 93.76\%). By increasing the number of representatives, our model would be significantly superior to LIME in all cases. In terms of the difference in explainability methods between TRUST and LIME, we argue that our model produces more straightforward explanations which are easier to understand. LIME is content with local approximations of the primary AI with other simple AI or ML models as a way of explainability and interpretability. Whereas, we model the behavior of the primary AI’s output with statistical measures. As statistics are more commonly a part of one’s educational background, they are much easier to understand compared to grasping the ideas behind the internal rationale of an AI or ML model.

\begin{table*}
 \centering
\caption{Log-likelihood}
   \includegraphics[scale=0.26]{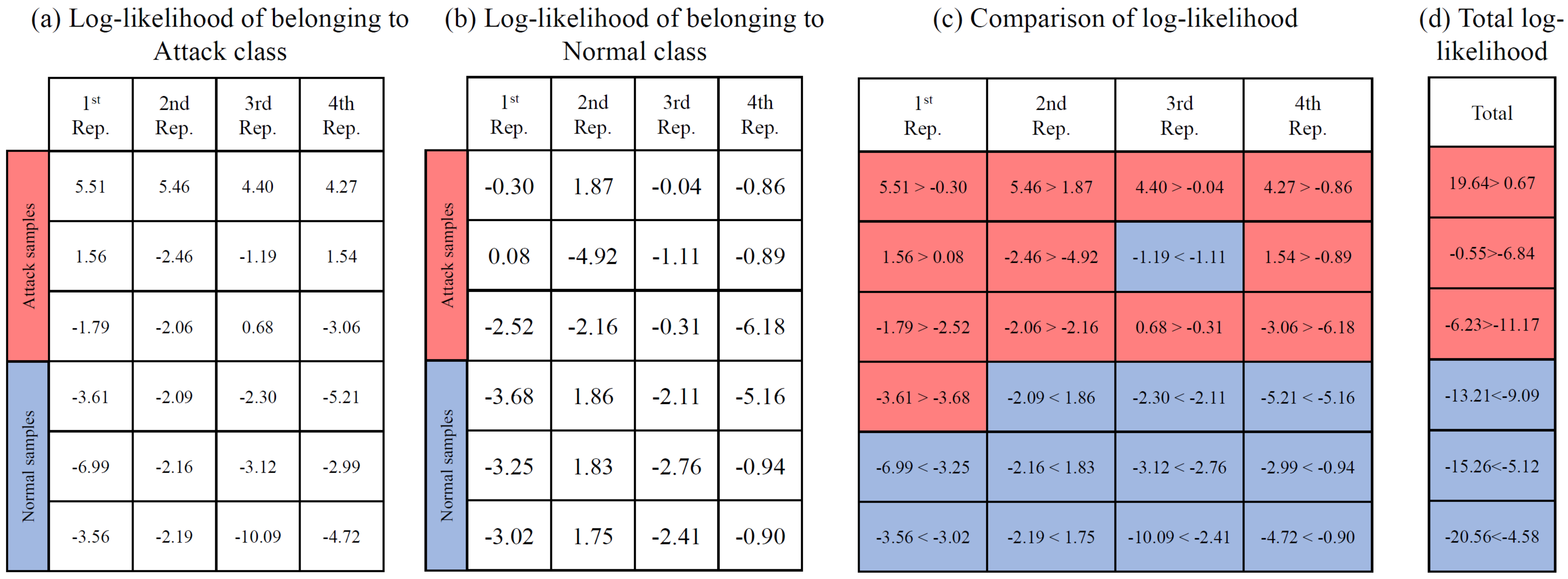}
  \label{log_result}
\end{table*} 

\section{How TRUST Explains The Primary AI Results to The User} \label{explaintouser}   

This section discusses how the TRUST explainer can explain the labels to a human user simply and interpretably on the NSL-KDD dataset as an example. As mentioned before, it is safe to assume that the user of the AI-based security system has a basic knowledge of AI, mathematics, and statistics. Each traffic packet in the NSL-KDD dataset is a 40-dimensional vector, but the TRUST explainer would calculate the representatives and reduce each sample to a four-dimensional vector. This process of dimension reduction is simply explained as follows.

The TRUST explainer modifies the feature values to remove redundancy in the data for a more transparent statistical analysis. By factorizing and removing their interdependencies and projecting them onto orthogonal dimensions, TRUST removes any correlation in the data. The transformed feature values are the factors introduced before. Then, utilizing a correlation metric (here, mutual information), the most revealing factors determining the class labels are chosen as representatives. Afterward, TRUST models the statistical behavior of the representatives and estimates their density functions with unique multi-modal distributions.

\begin{figure}
 \centering
   \includegraphics[scale=0.5]{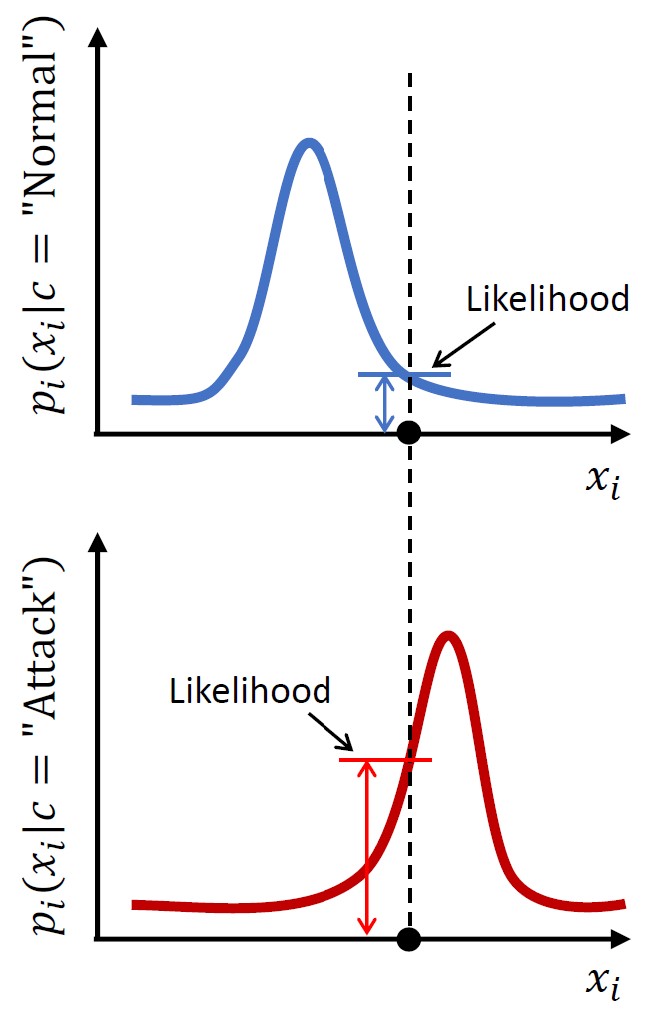}
\caption{An example of pdf curves for the $i$th representative}
  \label{likelihood}
\end{figure}

As mentioned before, statistics are more easily understood than the AI or ML models used as the XAI. For elaboration, we have shown an example in Figure \ref{likelihood}. It is trivial for the user to understand the likelihood differences of a new sample belonging to the pdfs of each class's representative. In this example, the likelihood of belonging to the attack class is higher.

To show the explanation process of TRUST, we randomly chose six samples (three attacks, three normals) from the test set. The log-likelihoods of belonging to each class using individual representatives based on Eq. \ref{log_likelihood} are computed. The results of each instance belonging to the attack class and belonging to the normal class are shown in Tables \ref{log_result}a and \ref{log_result}b, respectively. After an element-wise comparison of the values in these two matrices, the class that has a higher likelihood is marked in Table \ref{log_result}c. Notice that not all the representatives give us the right class all the time. For example, the \nth{1} representative makes mistakes with the \nth{4} sample. Similarly, the  \nth{3} representative makes mistakes with the  \nth{2} sample. However, when all four representatives are used to calculate the overall log-likelihood, as shown in Table \ref{log_result}d, the labeled class is correctly explained for all six instances.

\section{Conclusion and Future Direction}
XAI has attracted extensive attention recent years and has been treated as a necessary trend toward the next generation of AI. However, there is a significant lack of research work in this domain, specifically for numerical applications (compared to image applications). Meanwhile, several critical applications such as cybersecurity and IIoT are highly dependent on numerical data. The available work in the literature falls short from several aspects, such as trade-offs among performance, speed, reusability, and complexity, as discussed in this paper. Our proposed TRUST XAI is a breakthrough in this domain in integrating transparency, being independent, and comparability (in a statistical aspect) into AI-based systems. Here, we take the lead to use statistical principles to build XAI models that can accurately estimate the distributions of the AI's outputs and calculate the likelihood of new samples belonging to each class. Moreover, we have proven the superiority of our model to a currently popular XAI model, LIME. Our results show that our model is 25 times faster and more accurate than LIME making our model a great candidate for real-time and critical applications.

Despite TRUST's superior performance, it has some limitations as well. Due to using information gain in picking the representatives, TRUST might overfit to the training set. This would lead to poor performance on unseen data. On the other hand, if the Gaussian assumption cannot be made or the probability distribution of data changes, the output of TRUST would not be reliable. Also, the assumption of samples being drawn independently is very important.

As a future direction, for the selection of representatives, we can take into account other metrics such as gini index. In addition, we can extend the model to consider other possible probability distributions (e.g., Poisson). Also, removing the independence assumption requires approximating the joint probabilities of each pair of features.

\section*{Acknowledgment}
This work has been supported under the grant ID NPRP10-0206-170360 funded by the Qatar National Research Fund (QNRF) and NSF grant CNS-1718929. The statements made herein are solely the responsibility of the authors.

\bibliographystyle{IEEEtran}
\bibliography{trustxai_MZ}

\end{document}